# Cycle-GAN for eye-tracking

Rakhmatulin Ildar, PhD
*South Ural State University, Department of Power Plants Networks and Systems, Chelyabinsk city, Russia, 454080*
ildar.o2010@yandex.ru

**Abstract**
This manuscript presents a not typical implementation of the cycle generative adversarial networks (Cycle-GAN) method for eye-tracking tasks. For the implementation of this purpose, we developed the program that accepts images with any expansion from a camera and determines and cut the contour of the face and eyes by OpenCV and dlib libraries and converts these images in a resolution of 400 by 300 pixels. These images we used for training Cycle-GAN model. Usually, the Cycle-GAN does not require marking, but in this case, we used labeling. Images for training manually with an error of no more than 1 pixel for 4000 images were marked. In all labeling images, we painted a pupil with an unnatural for eyes bright green color and trained the neural network model to replace the pupil of the eye with a circle that is painted bright green. At the model's output, we have an image where the pupil is painted in bright green. In the final, the position of the pupil (given color in the video stream) can be easily detected due to the function of the OpenCV library – inRange. The program code and custom dataset are presented in the public domain.
https://github.com/Ildaron/6.Cycle-GAN-custom_data
https://www.kaggle.com/ildaron/dataset-eyetracking

## 1. Introduction

Eye-tracking (ET) is the process of determining the coordinates of gaze with a device used to determine the orientation of the optical axis of the eyeball in space. Earlier ET was mainly used in studies of psychophysics or cognitive development, but in the last decade advances in machine vision have allowed this technology to go far beyond the scope of disease diagnosis. Today eye-tracking is used to support multimedia learning, help in browsing the web, and is widely used in real-time graphics systems, which is especially popular in video games. But despite this, the main problem of modern eye-tracking systems is their high price. Equipment with accuracy of 2° has prices from several thousand dollars. Typically, these devices use the method of reflection of the cornea. The eyes are exposed to direct invisible infrared (IR) light, which results in the reflection in the cornea. The detection of this reflection is carried out by various functions of machine vision.

Our task is to use a webcam for eye tracking because this camera is available on almost every laptop or desktop computer. Therefore, it is advisable to shift the focus of research towards the development of neural network models that will detect eye using an image with low resolution from any webcams. In recent years, in the field of eye-tracking, the papers in which neural networks are involved have significantly prevailed. Deep neural networks implemented based on a convolutional neural network (CNN). The advantages of deep neural networks in eye-tracking tasks are detailed in the works [1,2].

There are a large number of works in which the standard use of popular models of deep neural networks is involved (OverFeat, Faster R-CNN, DeepID-Net, R-FCN, ION, MultiPathNet, NoC, G-RMI, TDM, SSD, DSSD, YOLOv3,4,5, FPN, RetinaNet, DCN, SSD300, SSD500, Tiny YOLO) where the speed of the model is determined by the structure, depth of the model and hardware [3,4]. Tracking accuracy in these works is several degrees [18, 19].

For us, the following research are interesting, in which the authors to improve accuracy showed creativity in using neural networks in eye-tracking tasks. Huang, B., et al. [5] used the recurrence module to refine the orientation of the eyes using the initial shape of the eyes. Only this method has low tracking accuracy. Unlike them, Fen, X.et al. proposed the new gaze detection model with a combination of superpixel segmentation and eye-tracking data [6], but it is not easy to use this model with a standard webcam. Oszelik, E. et al. [7], explained that the effect of color-coding allows more accurate information about eye movement to be obtained. In our manuscript, we used this color-coding effect to verify the accuracy of the developed model. Huang, H., et al. [8] proposed a two-phase CNN training strategy to combine head posture and viewing angles. This CNN architecture can reduce refit when training eye-tracking models directly with a head pose. In our research, the head position and its effect on the eyes did not increase the accuracy of tracking the developed eye-tracking model. Frank, H. et al. [9] introduced a system of stroboscopic catadioptric eye tracking. The new approach for mobile ET, based on cameras with roller shutters and stroboscopic structured infrared lighting was described. This method has high accuracy but requires a long setup and can easily go astray since the program only tracks glare. Krafka, K. et al. [10] introduced research the closest in content to this manuscript, where submitted the GazeCapture software.

In our research, we use similar facial processing technology to detect the area with the eyes and use a convolutional neural network to track the eyes in this limited area. The difference between this paper and our manuscript, that we use a new technique for labeling Cycle-GAN model. The disadvantage of a low-cost eye-tracking system is the low tracking accuracy when using weights that are presented with the model. Moreover, these models usually not always work as expected when training on a custom dataset. In this connection, this paper presents the implementation of the Cycle-GAN model for eye-tracking tasks. The presented model combines high precision, simplicity, and versatility.

## 2. Cycle Generative Adversarial Networks

Generative Adversarial Networks (GAN) introduced by Ian Goodfellow and colleagues in 2014 [11], allows you to generate very artificial images indistinguishable from real ones. The lack of GAN is the difficulty in tracking accuracy since the images are not paired. Therefore, the solution to this problem is to some extent Cycle-GAN, which, since the initial image is converted to the necessary photo and then back, allows you to check pixel by pixel the accuracy of the model.

The full explanation of the model is presented in the original article first published in 2018 [12]. The architecture of the model consists of two models of generators, where 1-Generator to generate images from 1-Dataset to 2-Dataset and 2-Generator to generate this image back (that is, from 2-dataset to 1-dataset). Each generator has its own discriminator model, which is trying to recognize whether the image is true. Due to the loss function, generators with each cycle improve their skills in tricking the discriminator, and the discriminator improves their skills to better detect fake images.

Unlike GAN, the number of manuscripts in which Cycle-GAN is involved is much less, so the facets of using this model still yet have not fully studied. Most of the research that uses the cycle-GAN is not related to our manuscript since in such works there is a conversion of images from one dataset to another, for example, conversion of geographical maps and photographs, etc. In our manuscript, the main target is the transformations in the video stream, the following works are remotely like our manuscript. Jin, X. et al. [13] presented an interesting case of the transfer of style in the video stream when facial expressions and attributes of one person can be completely transformed into another person. Kwon, Y. et al. [14] proposed an implementation of a model for predicting accurate and time-consistent future frames over time. But in our research more practical task - to find and replace only one object.

## 3. Prepare dataset

The use of Haar cascades and the dlib library directly to monitor the gaze position showed low and unstable results. But at the same time, these libraries can be used as auxiliary methods for finding the eye area. Working only with the eye area will significantly speed up and improve the accuracy of the model Cycle-GAN.

The Haar Cascade is a machine-learning method for detecting objects in an image, the idea of which was proposed in an article authored by Paul Viola and Michael Jones. A trained Haar cascade, taking an image as an input, determines whether it contains the desired object, i.e. performs the classification task. We used this method to determine the face in the video stream.

The landmark detection algorithm proposed by Dlib is an implementation of the Regression Tree Ensemble (ERT), introduced in 2014 by Casemi and Sullivan. This method uses a simple and quick function to directly estimate the location of a landmark. These estimated positions are subsequently refined using an iterative process performed by a cascade of regressors. Regressors make a new estimate from the previous one, trying to reduce the alignment error of the estimated points at each iteration. This method we used to determine facial contours. Figure 1 shows the implementation of these methods.

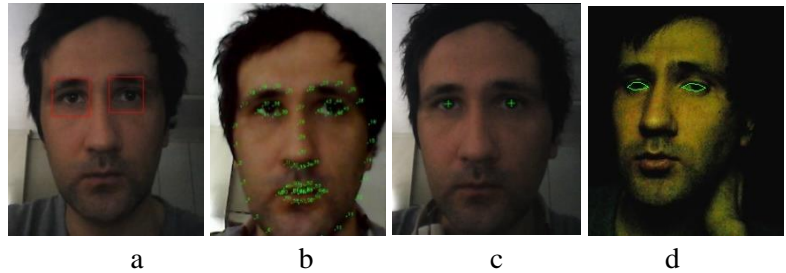

      a          b          c          d

Fig.1. Eye position search: a -haar cascade, b,c,d - various library implementation of dlib library

With dlib.shape_predictor command ("shape_predictor_68_face_landmarks.dat") we defined facial features, where shape_predictor_68_face_landmarks.dat is a trained model for 68 landmarks. We only take points 36 to 41 and add 30 pixels each point to expand the range, and then using the OpenCV ROI we limited the area with the eye in the video stream, this action is presented in figure 2.

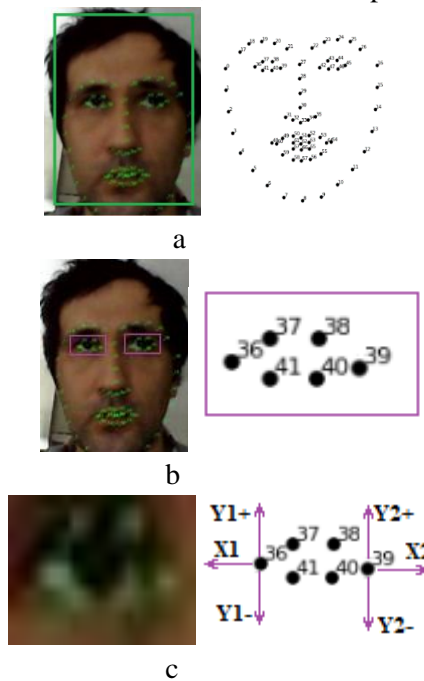

a

b

c

Fig.2. Detect eye in the video stream: a - face detection by OpenCV and face countors by dlib library, b – new video stream with points 36 to 41, c – expand video stream with points 36 to 41

Images from the video with the eye limited to rectangle and resolution 400x300 (fig.2.c) to train the model was used. We labeled 4000 images in which the model showed unstable results. As a result, it was decided to use an external dataset for the model pre-train, and use a custom dataset to train the last layers of the model.
We examined the following datasets can be used to train the neural network in the eye-tracking task:
- Columbia Gaze Data Set. Data set consists of 5,880 images of 56 people over varying gaze directions and head poses. For each subject, there are 5 head poses and 21 gaze directions per head pose (http://www.cs.columbia.edu/CAVE/databases/columbia_gaze/). There are several studies with this dataset [15];
- Openeds facebook dataset. Semantic segmentation data set collected with 152 participants of 12,759 images with annotations at a resolution of 400×640. Challenge participation deadline: September 15,

2019. But Dataset is still available on request. (https://research.fb.com/programs/openeds-challenge/) There are several studies on this dataset [16];

- MPIIGaze dataset. This data set consists of images taken in everyday conditions using the laptop's built-in webcams, in which 15 people participate. MPIIGaze dataset that contains 213,659 images (Xucong, Yusuke, & Mario, 2015);

- Image.net. massive database of annotated images designed to test and test methods of pattern recognition and machine vision.

- Kaggle dataset, Kaggle competition. Competitions are held periodically, participation in which open access to the dataset. (https://www.kaggle.com/c/gl-eye-tracking);

For pre-trainded Cycle-GAN we chose the following dataset (https://www.kaggle.com/4quant/eye-gaze), as it provides rather detailed annotation of the images. Subsequent training of the last layers will already be built on our own dataset.

As a result, we used 40,000 images in a 1280 x 720 pixels extension. Similarly to the personal dataset, we brought the image data to the following form, figure 3.

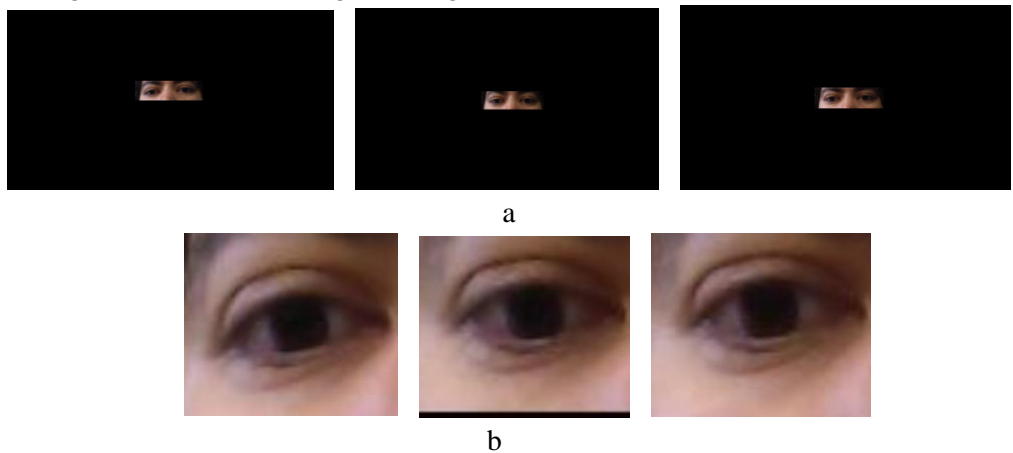

Fig. 3.a – examples images from the Kaggle dataset, b - converted images of the Kaggle dataset - 400 by 300 pixels

A personal dataset of 4000 images was manually marked. We received images during webcam operation (Defender G-lens 2597 HD720p, sensor resolution - 2 megapixels, frame rate for 1280x720 - 30). Then, after image processing, the pupil in a bright green color for the eye area - RGB: 45,253.9 were painted. For label 40,000 images from the Kaggle dataset, we wrote a program that processes the image to the required size and then takes the coordinates of the pupil from the table and draws a green circle instead of the pupil in the corresponding image and re-saves the images. Figure 4 shows the images after manual labeling for a dataset from Kaggle - a, and for a custom dataset – b

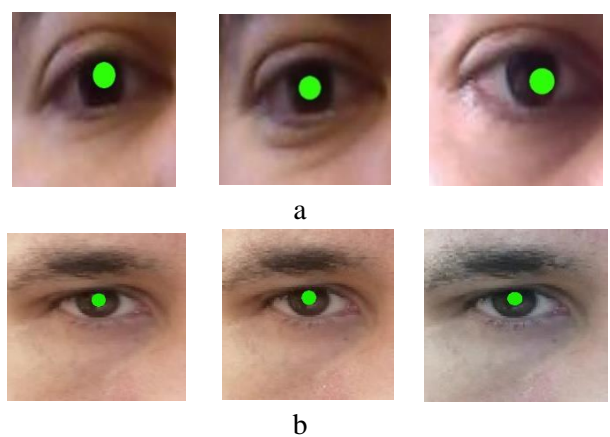

Fig.4. a – images from the Kaggle dataset after automatic labeling, b - images after manual labeling from the custom dataset

But there is an error in this Kaggle dataset, which is clearly visible in fig. 4, as a result of which the implemented program for drawing green circles instead of pupils has an error of several pixels. Therefore, we will use this dataset to obtain the initial weights. This method will allow us to train the model only in general terms - like choosing the right color and drawing a circle. That is why for the finish to train the model it is necessary to use a personal dataset.

## 4. Model implementation

Our model implementation, unlike the original model made on PyTorch, was performed on Keras. To implement the discriminator and generator models, we used the advice of the author of the Keras library F. Scholle which was given for the GAN model [17]:
- As the last activation function in the generator, we use tanh instead of sigmoid;
- We selected points from hidden space using the normal distribution (Gaussian distribution);
- We introduced a random component in two ways: using decimation in the discriminator and adding random noise to the discriminator labels;
- Instead of choosing the maximum value to reduce the resolution, we used alternating convolutions, and instead of the ReLU activation function, the LeakyReLU layer;
- To eliminate checkerboard artifacts caused by uneven coverage of the pixel space in the generator, we selected a kernel size that is a multiple of the step size for each use of the sparse Conv2DTranpose or Conv2D layers in the generator and discriminator.

The model code is given in the link in the annotation, description of models is available with the following commands: g_model_AtoB.summary(), g_model_BtoA.summary(), d_model_A.summary(), d_model_B.summary(), c_model_AtoB.summary(), c_model_BtoA.summary().

Discriminator trained models on real and generated images. Generator models are trained through discriminator models. These models are updated to minimize the losses predicted by the discriminator for the generated images. In this way, the generator generates images that fit the real dataset better.

Generator models are also updated depending on how efficient they are in regenerating the original image when used with another generator model, called loop loss. Finally, it is expected that the generator model will output an image without translation if an example is provided from the target area, called loss of identity. Each generator model is optimized through a combination of four loss functions: Match Loss, Loss of identity, Direct cycle loss, Reverse cycle loss. As the authors of the original article, we used the same loss functions, figure. 5.

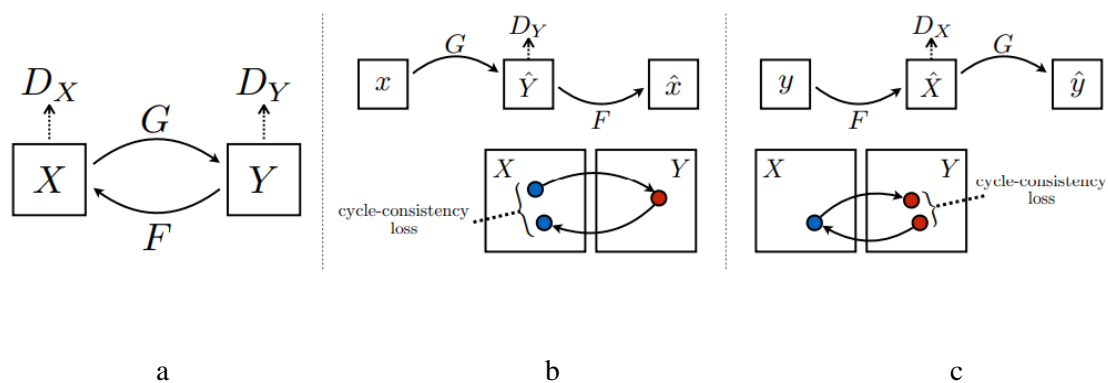

a  b  c

Fig. 5. Schematic representation of the Cycle GAN model. Where (a) Cycle GAN contains two mapping functions G : X → Y and F : Y → X, and associated adversarial discriminators DY and DX. DY encourages G to translate X into outputs indistinguishable from domain Y, and vice versa for DX and F. Two-cycle consistency losses that capture the intuition that if we translate from one domain to the other and back again we should arrive at where we started: (b) forward cycle-consistency loss: x → G(x) → F(G(x)) ≈ x, and (c) backward cycle-consistency loss: y → F(y) → G(F(y)) ≈ y

For the mapping function G: X → Y and it's discriminator DY we applied adversarial losses which are described by the following expression:

$$L_{gan}(G, D_Y, X, Y) = E_{y-p_{data(y)}}[LogD_Y(y)] + E_{x-p_{data(x)}}[\log(1 - D_Y(G(x)))], \quad (1)$$

where G is necessary in order to generate images G (x) similar to the dataset - Y. The discriminator DY seeks to distinguish between translated samples G (x) and real samples of the dataset Y. The task of G is to minimize this objective against opponent D who is trying to maximize it (minG, maxDY, LGAN(G, DY , X, Y )). A similar adversarial loss calculated for the mapping function F: Y → X and its discriminator $D_X$: that is, $\min_F, \max_{DX,} LGAN (F, D_X, Y, X)$.

A good result is achieved with direct and reverse generator losses of less than 0.5, an example of the learning process, figure. 6.

```
>2711, dA[0.120,0.170] dB[0.009,0.009] g[2.689,2.566]
for saved 101
>2712, dA[0.021,0.061] dB[0.004,0.016] g[2.104,2.002]
for saved 102
>2713, dA[0.110,0.145] dB[0.036,0.013] g[2.808,2.527]
for saved 103
>2714, dA[0.036,0.172] dB[0.020,0.012] g[2.232,1.841]
for saved 104
>2715, dA[0.142,0.105] dB[0.029,0.012] g[2.360,2.214]
for saved 105
>2716, dA[0.150,0.038] dB[0.015,0.012] g[3.169,3.150]
for saved 106
>2717, dA[0.035,0.140] dB[0.030,0.043] g[2.496,2.557]
for saved 107
>2718, dA[0.059,0.075] dB[0.017,0.070] g[2.501,2.334]
for saved 108
```

Fig. 6. Fragment of the model training

Figure 7-shows examples of images sent to the input of the model, and figure 7-b shows the images that we get at the output of the model.

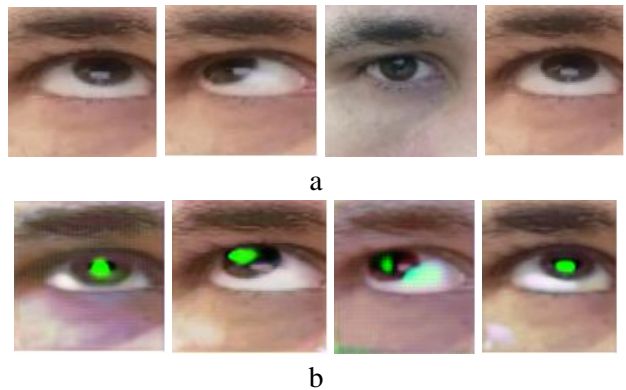

a

b

Fig.7. Examples of images sent to the model input, and b examples of images received by the model

Considering that a personal dataset was used for model training, we only labeled 4000 images. As a result, quite soon after 35 epochs, the model is over-trained. The retrained model on unfamiliar images shows the following result, figure. 8. (after 40 epoch).

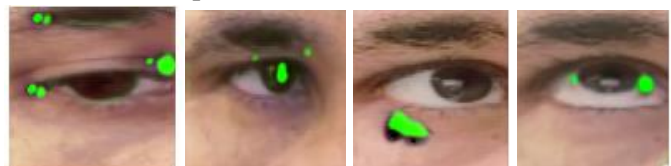

Fig. 8. Result retrained Cycle-GAN model

Training loss for both datasets represented in figure 9.

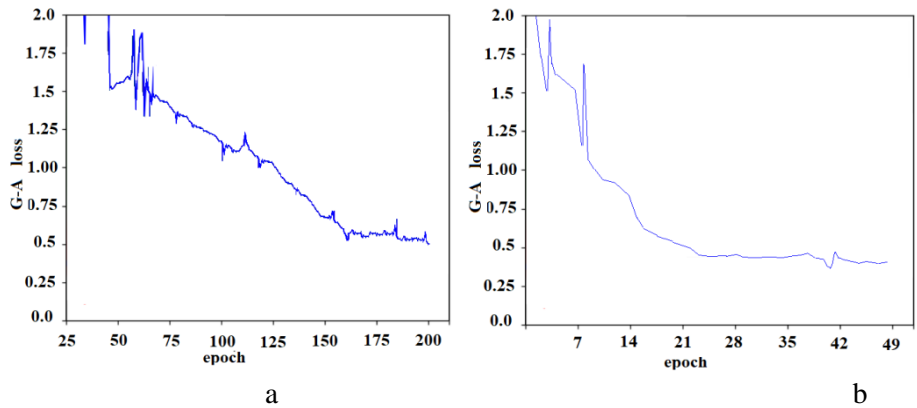

Fig. 9. Training losses for elementary training 40,000 images from the Kaggle dataset - a, and 4,000 images from the personal dataset - b

The initial training of the model requires a large number of eras. Training on a computer with a Gigabyte GeForce GTX 1060 WindForce -6G graphics card for one era was 45 minutes. Additional training of 4000 photographs for one era took about 4 minutes for one era and accuracy was achieved after 20 epochs. In the future, each user can retrain the model on a budget graphics card in just over an hour.

**5. Check model performance**

After receiving the image from the model cycle, we used the cv2.erode (image, kernel) functions to remove random green pixels, then, due to the cv2.dilating function, we restored the original image, after which the image took the following view, figure. 10.

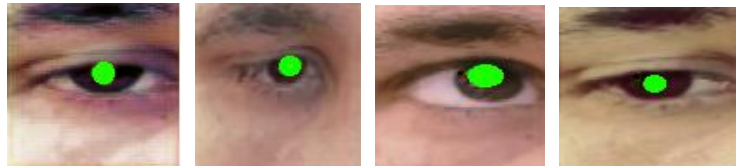

Fig. 10. Image obtained from the model Cycle-GAN and converted by the OpenCV library

The device was tested for conditions where the Haar cascades and the dlib library determined the position of the face and eyes with 100 % accuracy. Calibration of the device was carried out on 20 bright color squares with an area of 7.8 cm2 located on the screen of the monitor in the form of a desktop screen saver. The subject concentrated his attention on bright colored squares on the computer screen for 5 seconds (figure 11), the coordinates of each square were recorded in the computer's memory.

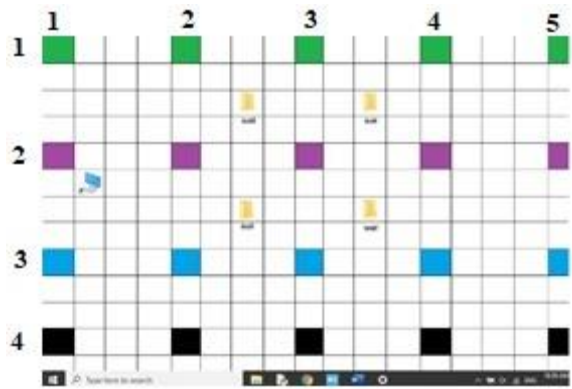

Fig. 11. Screen saver of the computer desktop for calibration and verification of the eye tracking device

The developed tracking device was checked by comparing the coordinates obtained during the tracking process with the coordinates obtained during the calibration process. The size of the monitor is 22 inches with a screen resolution of 1366 by 768 pixels. The object was 500 mm from the monitor in its natural position for work. The error of concentration of the subject's gaze on the monitor by one square gives an error of 2 degrees. The results are presented in Table 1. For each point, the average error for 30 measurements is given.

Table 1. Error in degrees for various sections on a computer monitor

| Error in degrees for various sections, 0 | | | | | |
|---|---|---|---|---|---|
| Position | 1 | 2 | 3 | 4 | 5 |
| 1 | 1 | 2 | 1 | 1 | 2 |
| 2 | 1 | 2 | 1 | 2 | 2 |
| 3 | 1 | 2 | 2 | 2 | 3 |
| 4 | 2 | 1 | 2 | 2 | 2 |

Total error is 1.7 degrees.

**Conclusion**

The manuscript describes and proves in detail that the implementation of the Cycle-GAN method is suitable for eye-tracking tasks. Usually, the Cycle-GAN cycle does not require marking, but in this case, we mark the image. Images for training were marked manually with an error of not more than 1 pixel. For this, the pupil was painted in bright green. Accordingly, the model learns to replace the pupil with a circle that is painted in bright green. After training Cycle-GAN and receiving weights, the image with the eye area obtained from the standard image of the webcam is fed to the model's input.

At the output of the model, we have an image in which the pupil is painted in bright green. In the end, the position of the pupil determined thanks to the function of the OpenCV library - inRange, which only needs to find the given color in the process of studying the color shade. The model was pre-trained on the Kaggle dataset. At the last stage, a model was prepared with frozen initial layers on its own data set. The program code is available in open access.

**Discussion**

In the future, it is advisable to try to use this module with a large number of the image in datasets. It is advisable to compare between different network models: BiGAN, CoGAN, GAN, SimGAN, CycleGAN, pix2pix, Ground truth, CycleGAN-VC2, PatchGAN, CycleGAN-VC. We used the method in which the pupil was painted over. It is probably advisable to paint over the eye itself and the area around the eyes, thereby making the image receive 2 color images at the output. It is necessary to train to search for a pupil in an image without preliminary image processing by external libraries. The presented model accepts the image only with the limited area with eye, but in practice, the dlib and OpenCV libraries are not very effective at identifying the area with the eyes.

**Conflicts of Interest**: None
**Funding**: None
**Ethical Approval**: Not required